\documentclass[conference]{IEEEtran}

\usepackage{adjustbox}
\ifCLASSINFOpdf
\else
\fi
\usepackage{graphicx}
\usepackage{amssymb,mathtools}
\usepackage{subfig}
\usepackage{multirow}
\usepackage{hhline}
 \usepackage[most]{tcolorbox}
 \usepackage{hyperref}
\IEEEoverridecommandlockouts  

\title{FPGA based Parallelized Architecture of Efficient Graph based Image Segmentation Algorithm }

\author{Roopal Nahar$^{1}$, Akanksha Baranwal$^{1}$, K.Madhava Krishna$^{1}$

\thanks{$^{1}$ Robotics Research Center, IIIT-Hyderabad, India}%
\thanks{{\tt\small roopalnahar08@gmail.com}}%
\thanks{{\tt\small akankshabar@gmail.com}}%
\thanks{{\tt\small mkrishna@iiit.ac.in}}%
}

\begin{document}

%


 \maketitle

\begin{abstract} Efficient and real time segmentation of color images has a variety of importance in many fields of computer vision such as image compression, medical imaging, mapping and autonomous navigation. Being one of the most computationally expensive operation, it is usually done through software implementation using high-performance processors. In robotic systems, however, with the constrained platform dimensions and the need for portability, low power consumption and simultaneously the need for real time image segmentation, we envision hardware parallelism as the way forward to achieve higher acceleration.
Field-programmable gate arrays (FPGAs) are among the best suited for this task as they provide high computing power in a small physical area. They exceed the computing speed of software based implementations by breaking the paradigm of sequential execution and accomplishing more per clock cycle operations by enabling hardware level parallelization at an architectural level.
\par

In this paper, we propose three novel architectures of a well known Efficient Graph based Image Segmentation algorithm. These proposed implementations optimizes time and power consumption when compared to software implementations. The hybrid design proposed, has notable furtherance of acceleration capabilities delivering atleast 2X speed gain over other implementations, which henceforth allows real time image segmentation that can be deployed on Mobile Robotic systems.
\end{abstract} 


%
\IEEEpeerreviewmaketitle

\section{Introduction}

Image segmentation is a key component of robotic vision systems and is used widely in applications that entail superpixelling and unsupervised segmentation. As a consequence, many different approaches have been proposed in this area like clustering \cite{Swendsen}, region-based growing \cite{Shapiro}, graph cuts \cite{Graph cuts}\cite{Felzenszwalb},super-pixeling \cite{Super Pixel}. However, nowadays, various methods of deep learning like CNN \cite{Semantic CNN}, SegNet \cite{Segnet}, semantic image segmentation \cite{Semantic CNN} are commonly used.
\par FPGA (Field-programmable gate arrays) implementations find relevance in Robotics due to its small physical area, light weight and its high capability of delivering tightly packed, energy efficient designs. FPGA provides real time parallelization, gate level control of system architecture allowing control over minute details of the arithmetic design. They also provide an  opportunity to pipeline sequential processes.
FPGA implementations of Image processing descriptors \cite{IROS FPGA}, collision avoidance \cite{Roopak}, further certify, that use of FPGA is an ideal solution for robotic systems with constrained dimensions. Hence, we can exploit the hardware flexibility, parallelization, logical, electrical and physical advantages provided by FPGA. And consequently, will help to  achieve real time and energy efficient image segmentation as compared to the software and other hardware implementations.The
research work that has been done in the field of FPGA
based Image Segmentation finds its applications in robotics,
computer vision and medical imaging \cite{Normalized graph cut}\cite{FPGA/ASIC implementation architecture}\cite{Real-Time Image segmentation}.
\begin{figure}[t]
  \centering
  \minipage{0.23\textwidth}
{\includegraphics[width=\linewidth, height=2cm]{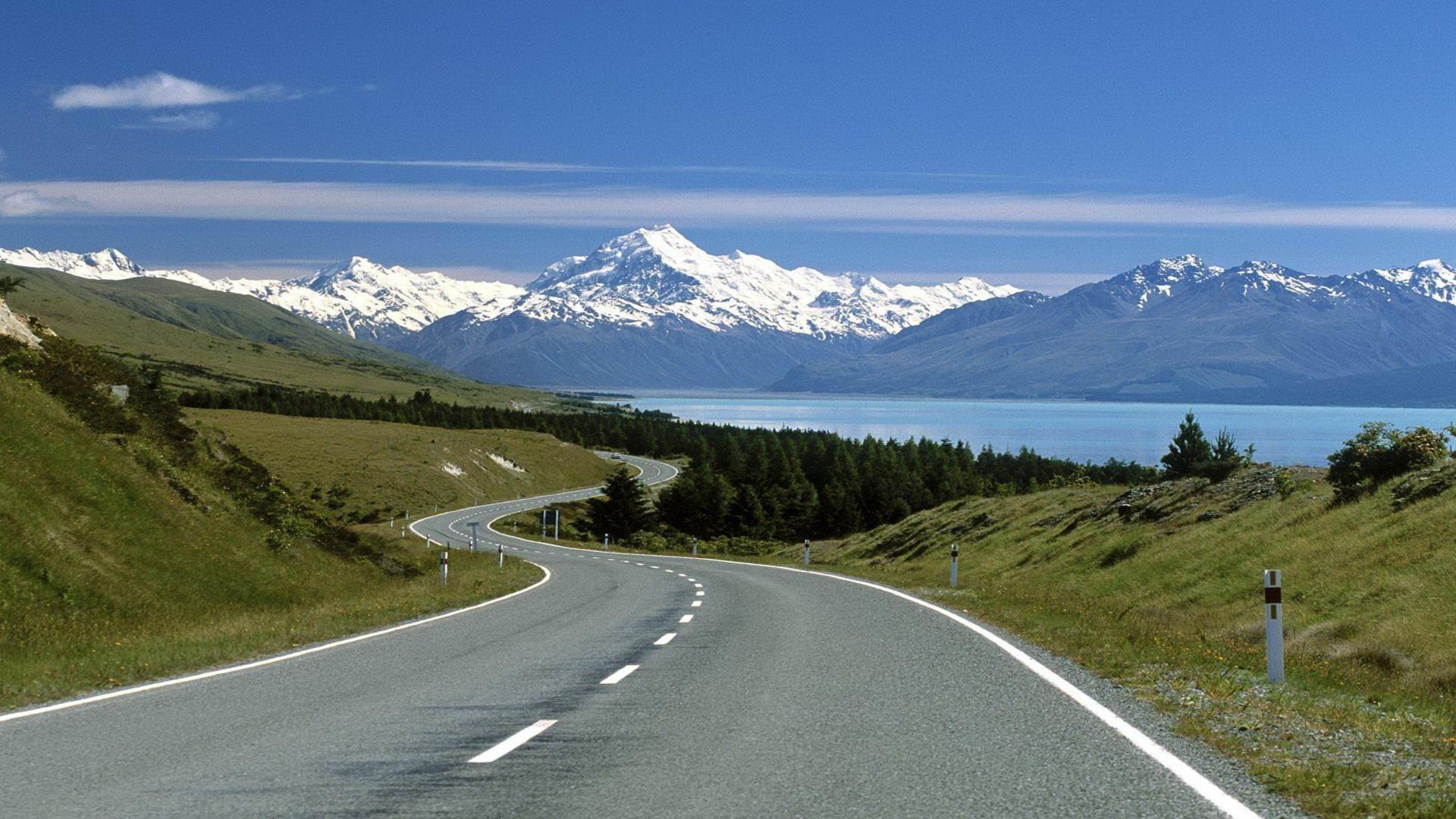}\label{fig:f1}}
  \endminipage  \hfill
  \minipage{0.23\textwidth}
{\includegraphics[width=\linewidth, height=2cm]{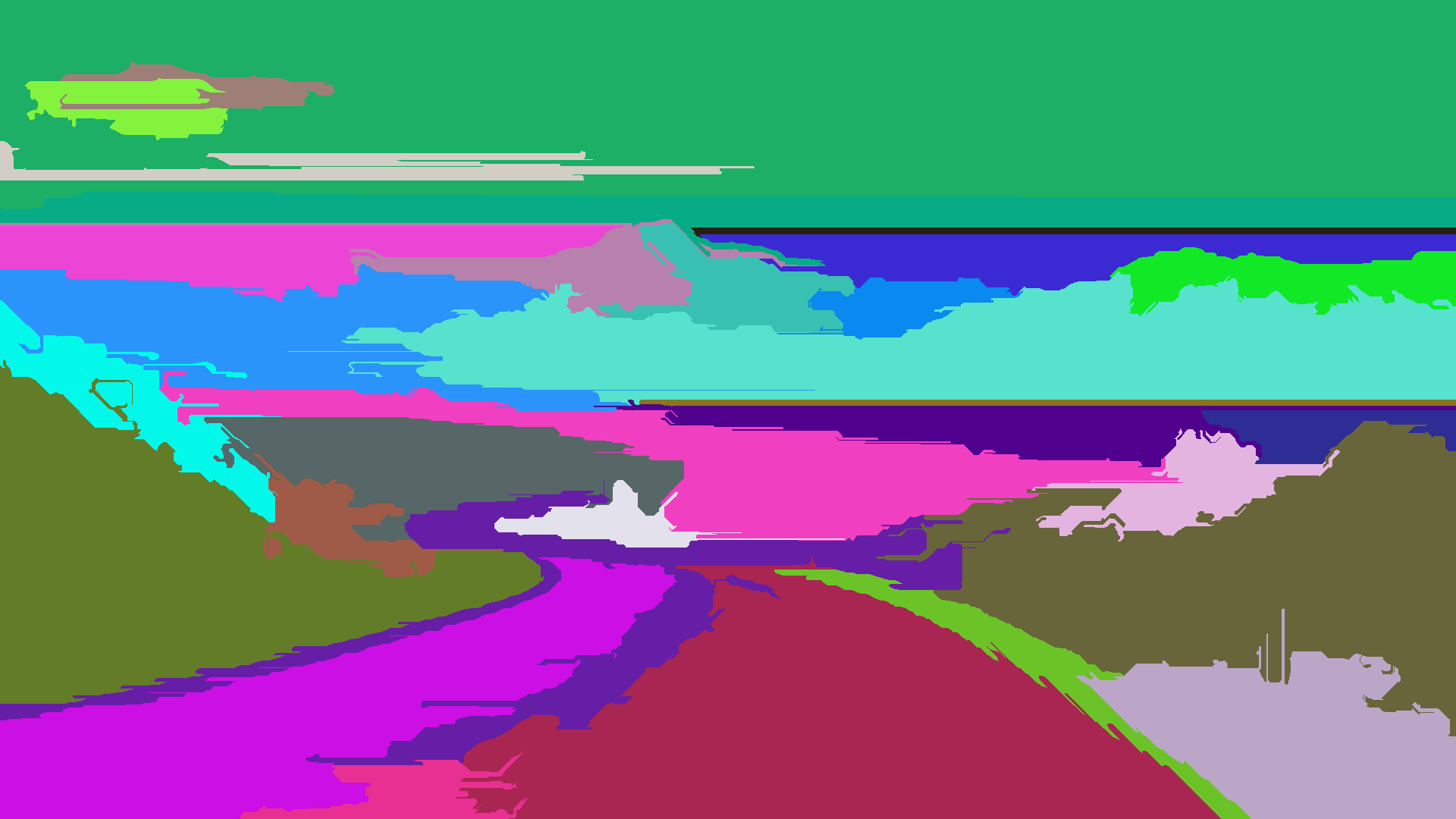}\label{fig:1}}
  \endminipage
  \caption{Input Image and Segmented Output Image using Efficient Graph Based Segmentation approach on FPGA}
\end{figure}
\begin{figure*}[t]
\centering
\minipage{0.85\textwidth}
  \includegraphics[width=\linewidth]{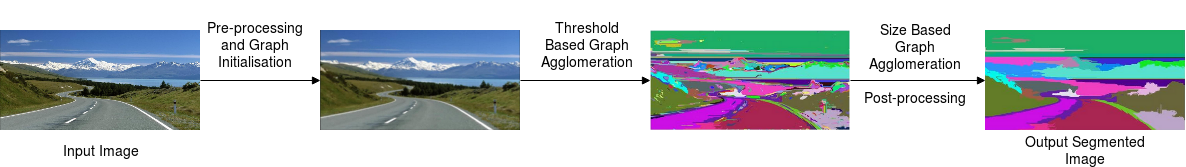}
\endminipage\hfill
\caption{Block diagram of Efficient Graph Based Image Segmentation.}\label{fig:2}
\end{figure*}
\par This paper focuses on three novel architectures -  Sequential, Pipelined and Hybrid of the well known Efficient Graph based Segmentation Algorithm\cite{Felzenszwalb} on FPGA. The sequential design exploits the gate level control of system architecture allowing the control over minute details of arithmetic design which is difficult in CPU implementation. Due to sequential nature of the sub-modules, this architecture is further modified to a pipelined design, by allowing interleaving of the processing steps of the sub-modules used in the algorithm. Parallelism and pipelining have been incorporated into the hybrid design by making multiple copies of elementary modules and using them in parallel, along with scheduling them efficiently. These hardware implementations reduce power dissipation and achieve real time segmentation. \par Comparative study of these three novel architectures has been done in terms of clock cycles and power dissipation to bring out vividly the advantages of FPGA over software implementation which generally follows sequential approach. The sequential approach shows a tangible improvement in computation time when compared to CPU implementation, but the hybrid architecture delivers an acceleration gain of at least 2X to the CPU implementation with much lesser power dissipation. The other results for the same will be delineated in more detail in the results section. The implementation is done using Verilog HDL language and the same is simulated and synthesized using  Xilinx Vivado Design Suite.


\section{Efficient Graph Based Image Segmentation Algorithm}
Efficient Graph based Segmentation algorithm by Felzenszwalb and Huttenlocher \cite{Felzenszwalb} has turned out to be popular due to its simplicity and high fidelity outputs.  An important characteristic of this algorithm is its ability to preserve detail in low variability image regions while ignoring detail in high variability regions. The overall flow of the algorithm \cite{Felzenszwalb} can be well-explained by the block diagram illustrated in Fig. 2.
\begin{figure}[h]
\centering
\minipage{0.18\textwidth}
  \includegraphics[width=\linewidth, height = 2.5 cm]{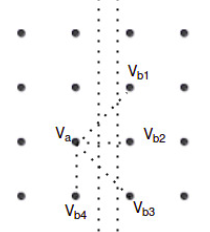}
\endminipage\hfill
\caption{\textit{Instead of considering the 8 conventional neighbors around a vertex V\textsubscript{a}, we consider only these 4 neighbors (V\textsubscript{b1}, V\textsubscript{b2}, V\textsubscript{b3}, V\textsubscript{b4}) for edge weight calculations to remove redundancy}}\label{fig:3}
\end{figure}
\par In this algorithm, a color image is given as input in RGB format which undergoes smoothening. Image is then represented as a weighted undirected graph \textit{ G=(V, E)}. Here,\textit{ V} represents vertices (the set of pixels in the image) and \textit{E} represents the set of edges defined between two adjacent vertices. For every vertex say V\textsubscript{a}, there will be four vertices (V\textsubscript{b1}, V\textsubscript{b2}, V\textsubscript{b3} and V\textsubscript{b4}) that will be considered for graph formation as shown in figure3. Each edge \textit{(v\textsubscript{a},v\textsubscript{b}) $\in$ E }has a corresponding weight \textit{ w(v\textsubscript{a},v\textsubscript{b})}, which is a non-negative measure of the dissimilarity between neighboring elements \textit{v\textsubscript{a}} and \textit{v\textsubscript{b}}. The weight of the edge \textit{E(v\textsubscript{a},v\textsubscript{b})} is the Euclidean distance between them in RGB color space. Preprocessing and graph initialization is done in the sub-module named Preprocessing and Graph initialization.

\par
In graph based approach, Segmentation \textit{S} is partitioning of \textit{V} into components such that each component \textit {C}$\in$ \textit{S}  corresponds to a connected component in a graph \textit{G'= (V,E')}, where \textit{E'} $\in$ \textit{E}. In Threshold based Graph agglomeration sub-module, we define the internal difference of a component \textit{(Int(C)) $C\subseteq $V}  to be the largest weight in the minimum spanning tree of the component, \textit{MST(C,E)}.\\
\begin{equation}
 Int(C) = \max_{e\in MST(C,E)}w(e)   
\end{equation}
       

We iterate over each edge to evaluate if there is any evidence of boundary between a pair of components \textit{(C\textsubscript{i}, C\textsubscript{j})} joined by the edge. Pairwise comparison predicate (\textit{D(C\textsubscript{i},
C\textsubscript{j})})is used to verify that if the two components are disjoint and the weight of the edge joining them is less than the minimum internal difference (\textit{MInt}) of both the components then they are merged on basis of threshold function $\tau$ using Eq. 2.
Pairwise comparison predicate \textit{D(C\textsubscript{i},
C\textsubscript{j})} is defined as
\\
\begin{equation}
D(C\textsubscript{i}, C\textsubscript{j}) = 
\begin{cases}
    true & \text{if Dif(C\textsubscript{i}, C\textsubscript{j})$>$\textit{MInt}(C\textsubscript{i}, C\textsubscript{j})} \\
   
    false & \text{otherwise}
  \end{cases}
\end{equation}
where the minimum internal difference, \textit{MInt}, is defined as,
\\
\begin{equation}
\textit{MInt}(C\textsubscript{i}, C\textsubscript{j}) 
=  min(\textit{Int}(C\textsubscript{i}) +     \tau (C\textsubscript{i} ),\textit{Int}(C\textsubscript{j}) + \tau(C\textsubscript{j})).
\end{equation}
and the difference between the two components \textit{Dif (C\textsubscript{i}, C\textsubscript{j})} is defined as:
\\
\begin{equation}
Dif(C\textsubscript{i},C\textsubscript{j})=   \min_{v\textsubscript{i}\in C\textsubscript{i}, v\textsubscript{j}\in C\textsubscript{i},(v\textsubscript{i},v\textsubscript{j})\in E}w(v\textsubscript{i}, v\textsubscript{j})   
\end{equation}
\begin{figure*}[t]
\centering
\minipage{0.9\textwidth}
  \includegraphics[width=\linewidth, height = 5.5 cm ]{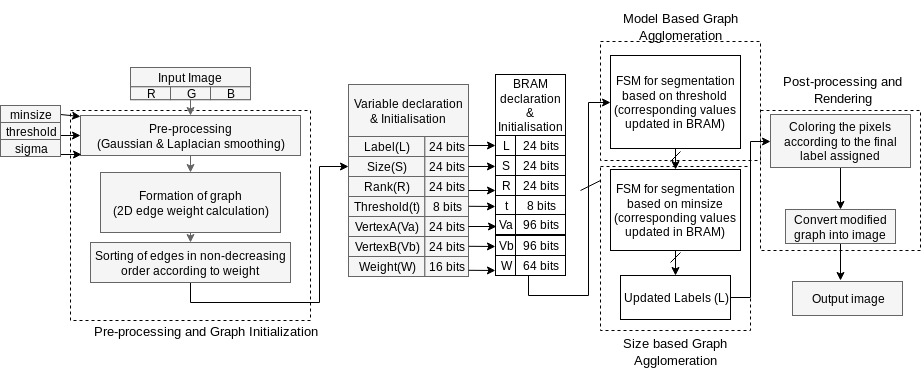}
\endminipage\hfill
\caption{Block diagram of proposed \textit{Sequential Architecture}}\label{fig:dp}
\end{figure*}
\par The threshold function $\tau$ used in Eq. 3 is used to control the degree, to which \textit{Dif(C\textsubscript{i}, C\textsubscript{j})}  must be larger than  \textit{MInt(C\textsubscript{i}, C\textsubscript{j})}. Threshold function $\tau$, is defined as,
\begin{equation}
\tau (C) = \textit{k}/|C| 
\end{equation}
where \textit {$|$C$|$} is the size of the component and \textit{k} is some constant.
\par Size based Graph Agglomeration sub-module does merging of the components based on min\_size factor. Components get merged with their neighboring components if their sizes are less than min\_size which is defined by the user. Post-processing and rendering sub-module reconstructs the modified graph into an image. It also recolors the image based on the new labels assigned to the pixels. This new image formed is the segmented output.\\

\section { Proposed Architectures for FPGA }

Image segmentation using graph based approach involves solving equations [1-5] for each component of the graph and finally merging them based on threshold and min\_size. This section delineates the three proposed architectures  - \textit{sequential, pipelined} and \textit{hybrid} in detail below: 

\subsection{Sequential Architecture}
Figure 4 shows the architectural flow of the sequential implementation. 
After the initial preprocessing and graph initialization, as explained in the previous section, the sorted values of \textit { V\textsubscript{a}, V\textsubscript{b}} and \textit{W} are stored in three different Dual Port BRAMs. Due to dimensional constraints, while storing in BRAM, each address of BRAM corresponds to four data segments. Every vertex has four attributes namely rank, label, threshold and size. All labels are assigned a different value and size 1 because initially each vertex is considered as a different component. Threshold for all vertices is assigned the same value as that of the global threshold. After variable declaration and initialization, these values are stored in four different Dual Port BRAMs.  These seven BRAMs are sent as input to Finite State Machine which updates the four attributes based on threshold criterion as explained in Eq. 1-5. An analogous Finite State Machine is used for merging segmented components based on min\_size. The modified graph obtained is then reconstructed into an image. Random color assignment is done to the pixels, based on final updated labels.
\par Sub-modules used to implement this algorithm are as follows:
\begin{itemize}
  \item \textbf{Finite State Machine for Segmentation:} 
  The Finite State Machine module(Fig. 5) is used for implementing the blocks - threshold based graph agglomeration and size based graph agglomeration. As shown in Fig. 4, seven BRAMs are sent as input to the block Threshold Based Graph Agglomeration. Depending on the value of address, we read values from V\textsubscript{a}, V\textsubscript{b} and W BRAMs. Each V\textsubscript{a}, V\textsubscript{b} and W corresponds to   the undirected weighted edge between them.  As soon as V\textsubscript{a}, V\textsubscript{b} are read, find modules are used to obtain their corresponding parent labels.
  \begin{figure}[h]
\centering
\minipage{0.45\textwidth}
  \includegraphics[width=\linewidth,height = 3.5cm]{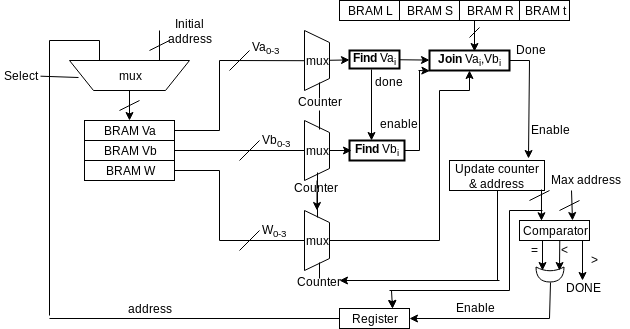}
  \endminipage\hfill
\caption{Finite State Machine Architecture}\label{fig:f}
\end{figure}
\par These two parent labels representing components are given as an input to the join module which decides whether to merge the components or not based on threshold criterion (Eq. 2). Find and Join modules are explained in the sub-section below. When the join module asserts the signal done, counter and address are updated as per the requirement.  Once all the edges have been traversed the state machine is terminated. This state machine updates the four attributes - rank, label, size and threshold. These updated BRAMs are again sent as input to a similar Finite State Machine which merges components based on min\_size and re-updates those four attributes. Once all the edges have been traversed the state machine is terminated.\\
\item \textbf{Find module:}
The Find module (Fig.  6) is used to search the parent label of the current component. 
A component when merged with any other component, gets the new label using set union find algorithm[18]. This updated label is stored in the Label BRAM and further given as an output for later modules.
\begin{figure}[h]
\centering
\minipage{0.45\textwidth}
  \includegraphics[width=\linewidth,height = 3cm]{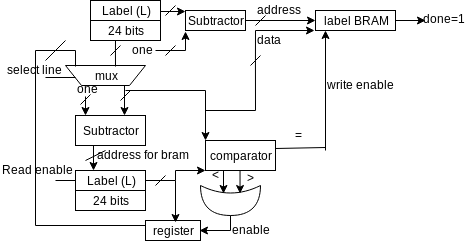}
\endminipage\hfill
\caption{\textit{FIND} Module Architecture}\label{fig:dp}
\end{figure}

\item \textbf{Join module:}
The Join module (Fig. 7) is used for merging the 2 components. This module is used in Threshold based Graph Agglomeration state machine (Fig. 4) and is used to merge components based on threshold criterion (Eq 2). In Size based graph agglomeration, this module is used for merging components based on min\_size criterion as explained in Section 2. Rank based set union algorithm[18] is used to implement this algorithm and reduce time complexity. 
\begin{figure}[h]
\centering
\minipage{0.45\textwidth}
  \includegraphics[width=\linewidth, height = 3.75 cm]{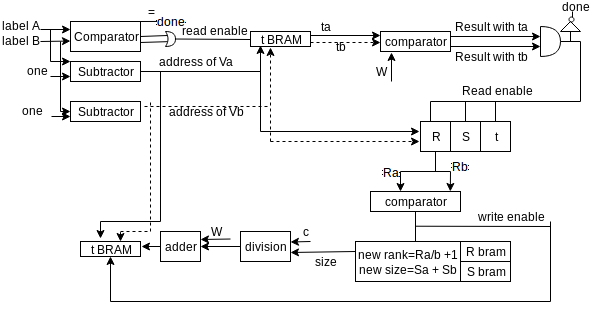}
\endminipage\hfill
\caption{\textit{JOIN} Module Architecture}\label{fig:dp}
\end{figure}
\end{itemize}
This sequential architecture uses these above modules as shown in Fig. 4. It exploits the gate level control provided by an FPGA and henceforth, allowing to control over minute details of the arithmetic design which is not possible in CPU implementation.
\subsection{Pipelined Architecture}
Figure 7 shows the process flow for the pipelined implementation of this algorithm.
In the fig. , all steps in one horizontal row are executed parallelly and as we move to next row, all previous tasks would be completed.
Due to sequential nature of the sub-modules, internal Pipeline processing is possible. While preprocessing of image and graph formation is being done, simultaneously initialization of the four BRAM can be done as these are mutually exclusive. Also, the edge weights can be computed row wise in parallel thus, graph formation time can be reduced to O(n) from O(n*n). Even for random recoloring, since colors depend only on the label assigned, similar reduction in time complexity is seen. Dual port BRAMs have two independent access ports. Exploiting this as soon as the first set of sorted Va, Vb and W are stored in BRAM, the FSM for segmentation based on threshold can start executing the first iteration. Using this also find and join operations can be pipelined with the write operation. This substantially reduces the time. Further finding the parent labels of Va and Vb are independent processes and can be executed together. Since there are different dual port BRAMs for label, rank and threshold,  find and updating threshold operations are pipelined. Similarly, read and join operations are pipelined. Independent dual port BRAMs also allow reading the labels before threshold based merging is finished. Thus, the last iteration of threshold based merging is pipelined with the first iteration of min size based merging.  Thus by making elegant modifications in the sequential architecture, it is possible to save a tangible amount of clock cycles. 
\begin{figure}[h]
\centering
\minipage{\columnwidth}
  \includegraphics[width=\linewidth, height = 9cm]{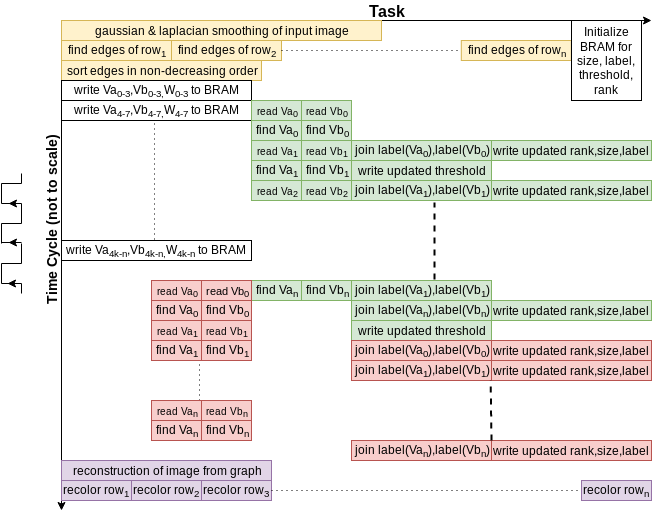}
\endminipage\hfill
\caption{Process flow of the pipelined architecture }\label{fig:dp}
\end{figure}
\subsection{Hybrid Architecture}
When there is no limit to the number of resources that can be used, Hybrid architecture can be designed as shown in Fig 9. This architecture is a full fledged parallel and pipelined implementation of the algorithm. By taking advantage of parallelism provided by an FPGA, we use multiple copies of sub-modules and use them in parallel.
\par An input image can be divided into n parts, and each part will then undergo independent pre-processing, Graph initialization and Threshold based Graph agglomeration.
Now, these parts are merged using Horizontal and Vertical stitching. This new stitched image undergoes size based graph agglomeration and random recolouring based on new labels. The value of n needs to be chosen judiciously. If n is very large then due to the limited number of resources on FPGA, this hybrid architecture will not work. Also, larger n implies smaller sub-images, which may result in loss of information while merging based on threshold and stitching. \par This flow is illustrated with an example below.
 \begin{figure}[h]
\centering
\minipage{0.42\textwidth}
  \includegraphics[width=\linewidth,height =  8cm]{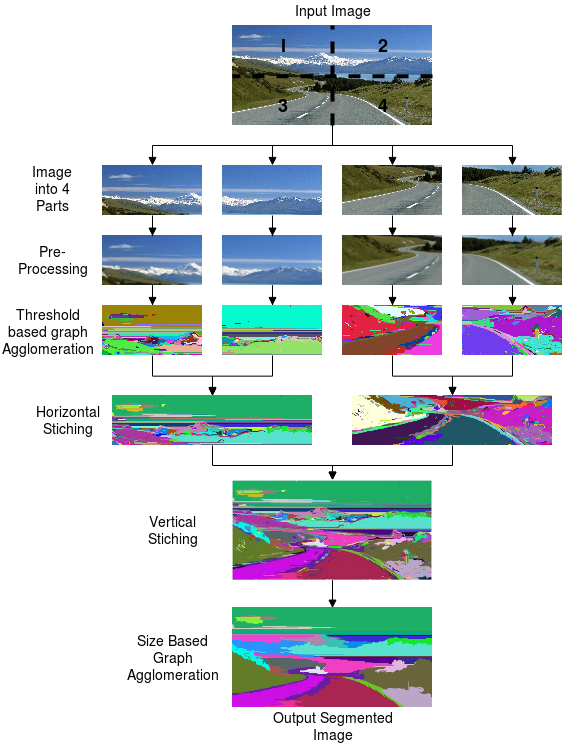}
\endminipage\hfill
\caption{Block diagram of proposed \textit{Hybrid} Architecture}\label{fig:dp}
\end{figure}
\par As shown in a fig. 9, the input image is divided in 4 parts. Each sub image undergoes pre-processing parallelly. This smoothened sub-images undergoes segmentation based on threshold(Eq. 2) independently. Now sub-images 1-2  and 3-4 are stitched horizontally.

 \par For horizontal stitching, the rightmost column of the left image is compared with the leftmost column of the image to the right. As the individual images have already been segmented according to the threshold, while merging along the joining edges, instead of considering four neighbors for edges, only one neighbor is considered. We consider  V\textsubscript{a} -V\textsubscript{b2} and ignore V\textsubscript{a} - V\textsubscript{b1}, V\textsubscript{a} - V\textsubscript{b3}. The dotted line in (Fig 3) shows the neighboring edges for horizontal stitching. This reduces a tangible amount of computation. Similarly, vertical stitching is implemented using the bottom most row of the top image and the topmost row of the bottom image. 
\par Now this image which we get after vertical stitching, is the segmented output which we get after thresholding. 
Segmentation based on min\_size is then done over the complete image. Random coloring of the image based on new labels results in the final segmented image.


\par This hybrid architecture saves a lot of clock cycles and helps in achieving real time segmented implementation.

\section{FPGA IMPLEMENTATION }
Optimized for high-performance logic and DSP with low
power serial connectivity, the design test platform, Virtex UltraScale XCVU190-2FLGC2104E FPGA delivers 4 CPF4 Optical Interfaces, 28 Gbps Backplane Interface, Dual 512MB Quad-SPI flash memory, 2GB HMC Memory to  meet higher bandwidth, performance and memory demands with less power. The table below provides system parameters.


The three architectures explained in the previous sections are implemented on this FPGA.The 7 BRAMs used in the design are simple dual port memories because they have two independent access ports to a common storage array. The modules Find and Join are implemented using finite state machines and various IP cores like Divider Generator are deployed for computation of division. 
\begin{table}[!htb]
\centering
\caption{Comparative study of resource utilization in the three architectures}
\begin{tabular}{|c|c|c|c|c|}
\hline
\textbf{Architectures} & \textbf{CLBs} & \textbf{LUTs} & \textbf{Power dissipation (mW)} \\
  \hline  

  \textbf{Sequential} & 1352 & 332 & 396 \\
  \hline
  \textbf{Pipelined} & 2426 & 525 & 478 \\
  \hline
  \textbf{Hybrid} & 4638 & 941 & 740  \\
  \hline
  \end{tabular}
\end{table}
To reduce number of resources, serial operation is used at certain places. The design of system comprises both serial and parallel operations to maintain an optimum level of utilization of resources as well as clock cycles required. The
exact values involved are presented in results section.

\section{Results}

The three architectures discussed above have been implemented on FPGA. We realized that if FPGA has lesser resources, then pipelined architecture should be preferred over the Hybrid one. On modern FPGA hardware, using Hybrid architecture, an image can be split into more sub parts (a large value of n say 16 or 32), exponential improvement in computation time can be observed.
The entire segmentation module - threshold based graph agglomeration and size based graph agglomeration runs on FPGA board, while the processor is used for pre-processing and rendering. 
We discuss the performance of the three architectures designed based on acceleration in computation and power dissipation in detail below.

\begin{figure*}[!tbh]
  \centering
  \minipage{0.30\textwidth}
  \subfloat[Input Road image1]{\includegraphics[width=\linewidth, height=2cm]{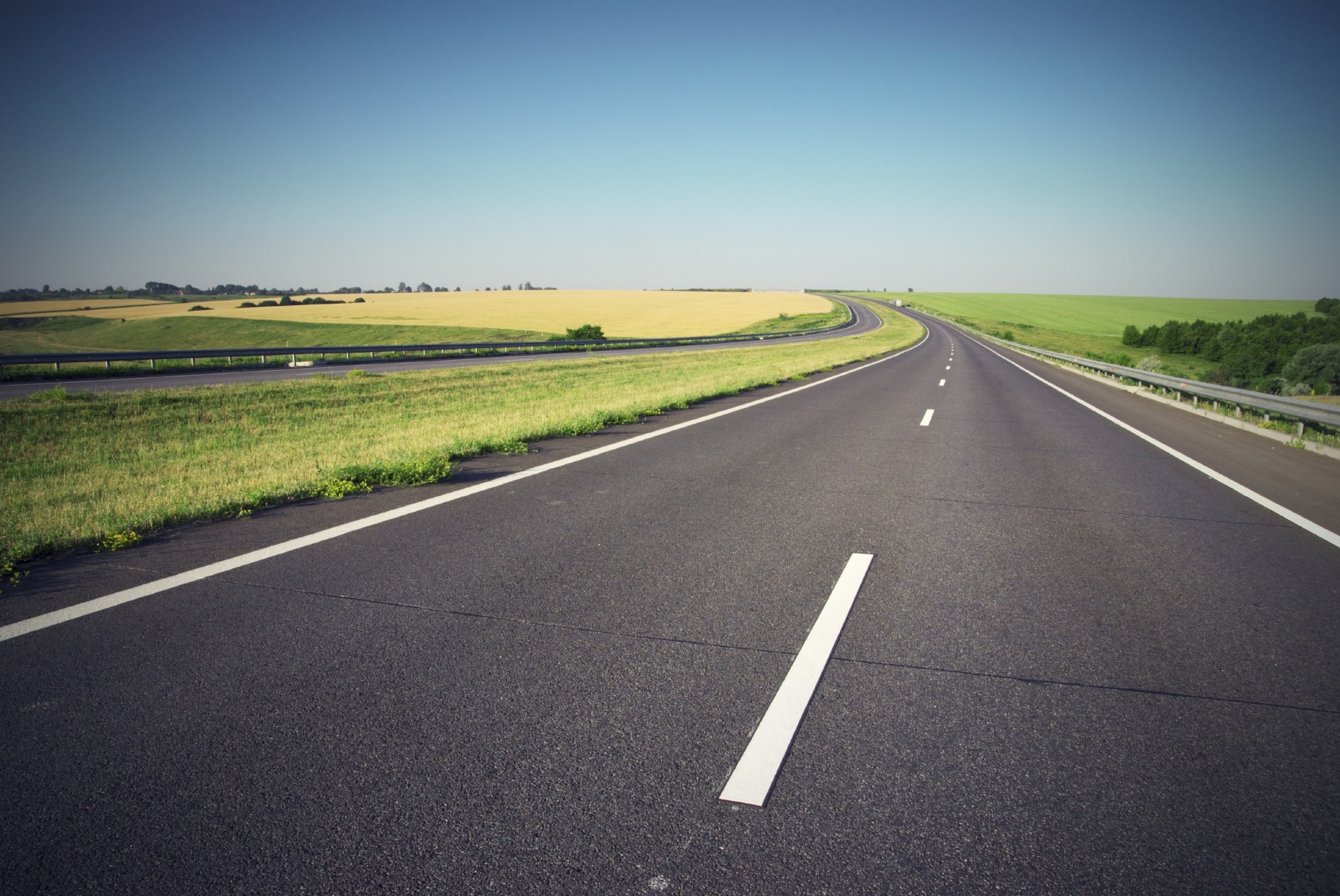}\label{fig:f1}}
  \endminipage  \hfill
  \minipage{0.30\textwidth}
  \subfloat[Segmented image after thresholding]{\includegraphics[width=\linewidth, height=2cm]{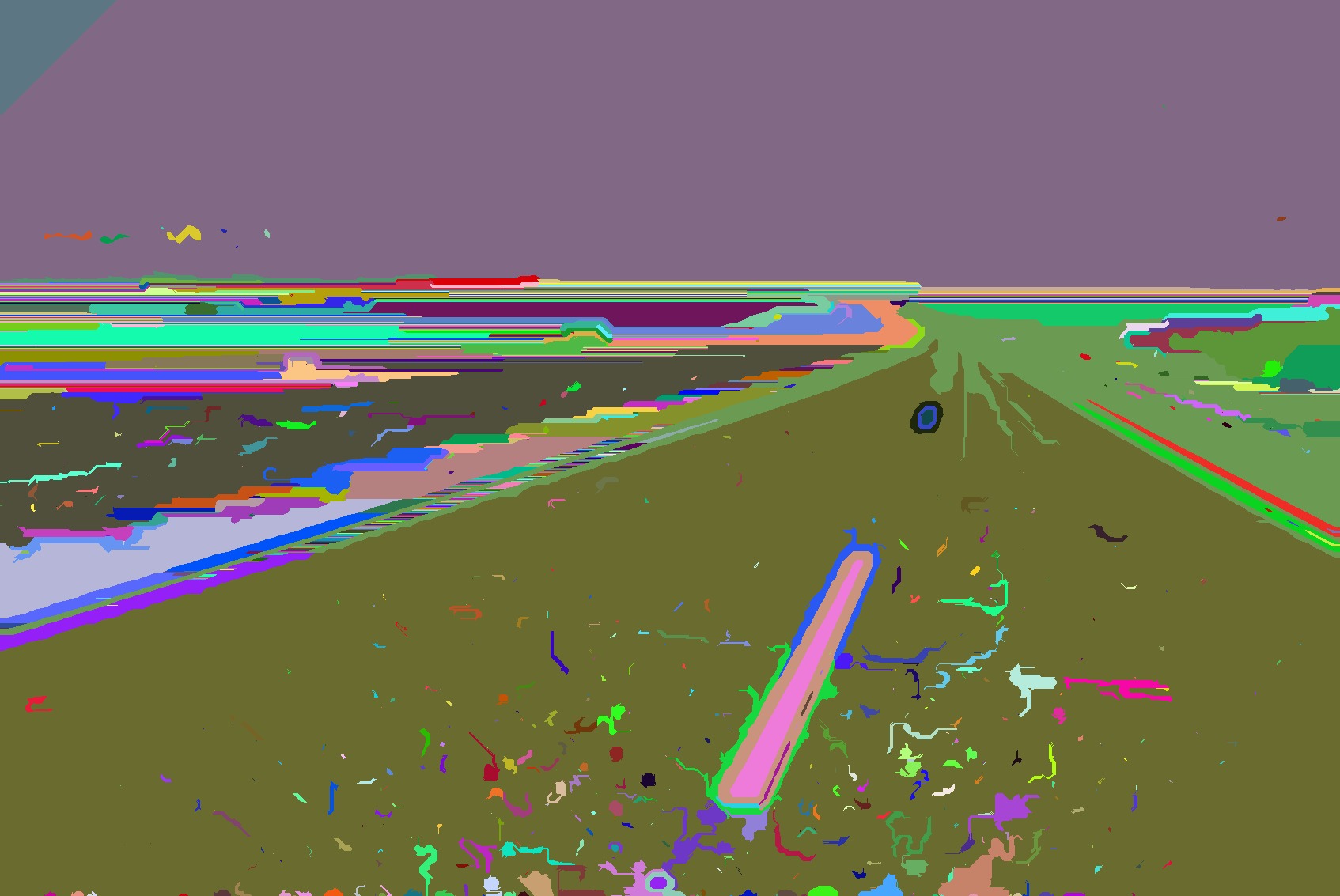}\label{fig:f2}}
  \endminipage\hfill
  \minipage{0.30\textwidth}
  \subfloat[Segmented image after min-size]{\includegraphics[width=\linewidth, height=2 cm]{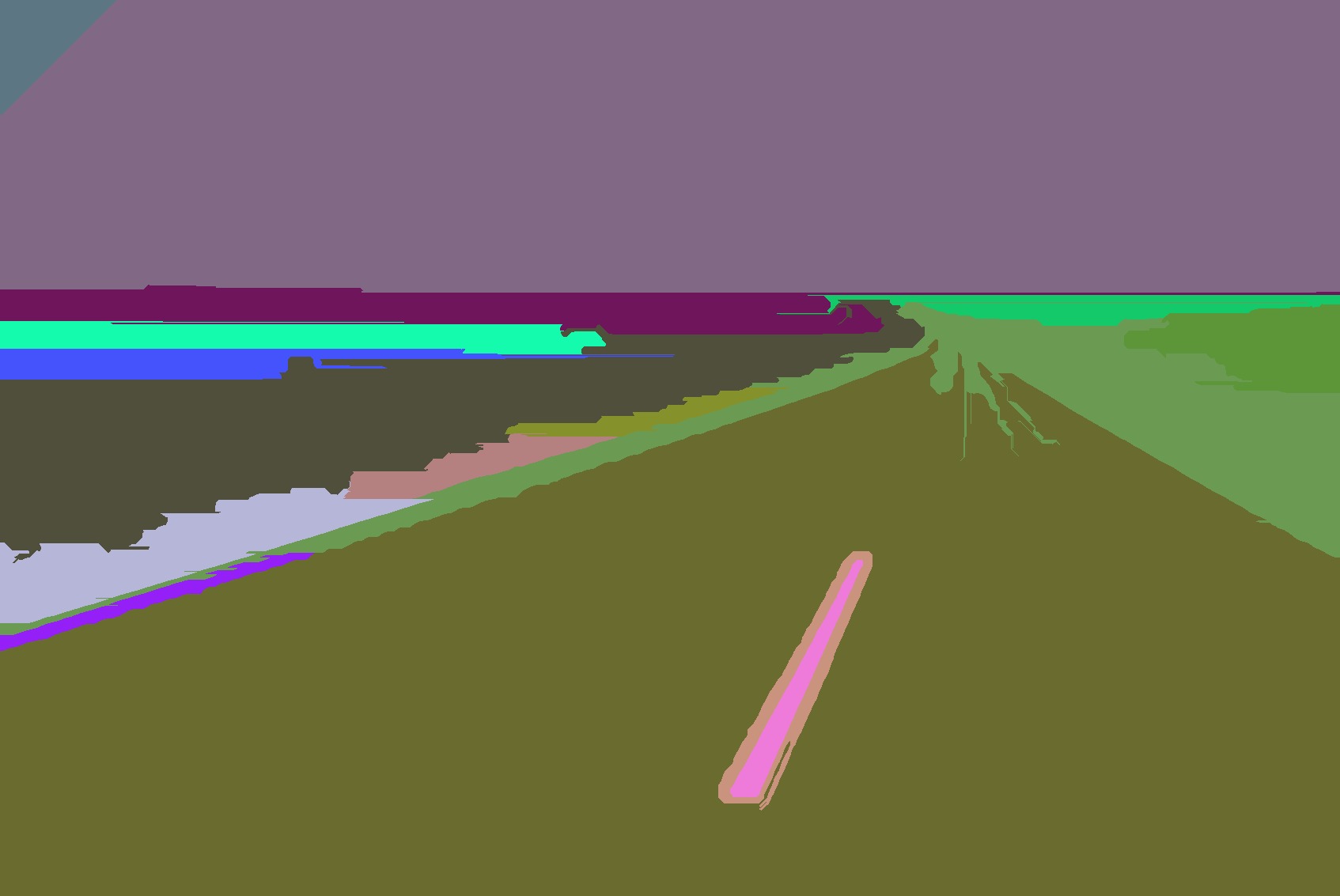}\label{fig:f3}}
  \endminipage  \hfill
  \caption{Input images (1920 X 1080 color image) and showing intermediate results and final segmentation using Efficient Graph Based Segmentation approach on FPGA }
\end{figure*}
\subsection{Acceleration in computation}
Table 3 shows the computation time a CPU takes and various other proposed architectures. We can see that a lot of clock cycles can be saved when we go for a hybrid architecture instead sequential or pipelined design. We can see that a significant amount of speed up is obtained, even though it has high resource utilization but this issue can be addressed by the ability of FPGA to dynamically reconfigure. At a particular instance, we can have only those resources which are required for processing
and FPGA can be reconfigured before the next step. In this way, highly parallel architectures can be designed on FPGA.
\par As we can observe from Table 3, the sequential architecture shows a marginal improvement compared to the CPU implementation. But by making the design more tightly packed and introducing pipelining at sub-modular level, we can see further more improvement in computation time. By breaking the paradigm of sequential
execution and accomplishing more per clock cycle operations by enabling hardware level parallelization at an architectural level , we can achieve enormous speed up.

We can infer from table IV, that as we keep on increasing the value of n, we achieve more and more acceleration. But, the value of n needs to be chosen judiciously. It depends on two factors- size of image and the FPGA board that is being used. For example, an image of size 128*72, if we split this image into 16 parts, the result obtained is very distorted.
Also, if n is very large then due to the limited number of resources on FPGA, this hybrid architecture will not work.

\subsection{Power consumption by equivalent hardware}
Simulation of power dissipation has also been done using Vivado Power analysis tool. The results are compiled in table V. As evident from the results shown, sequential implementation has lowest power dissipation and
hybrid implementation has highest. This is because, in the hybrid architecture, more cells and interconnects are active at any
instance of time compared to sequential. Even in pipelined architecture, the number of
active interconnects are more than that in sequential because the design is tightly packed. Hence power dissipation of these two architectures is more when compared to sequential.
\par The power dissipation in FPGA is in the order of milliwatt which is nominal when compared to power dissipation in a typical CISC or RISC processor for which power dissipation is in the order of Watt. Hence, both in terms of power dissipation and computation time, hybrid architecture is way better than the CPU implementation.
\begin{table}[t]
\centering
\caption{Computation time obtained for different architectures for different sizes of test images}
\resizebox{0.7\columnwidth}{!}{  
\begin{adjustbox}{max width=9 cm, max height = 3.5 cm}
\begin{tabular}{|c|c|c|c|c|}
\hline 
\multicolumn{4}{|c|}{\textbf{Computation time (in ms)}} \\
\hline
\textbf{Image Size} & \textbf{128 X 72} & \textbf{256 X 144} & \textbf{512 X 288} \\
  \hline  
  \textbf{CPU} & 49.2 & 96.31 & 261.14 \\
\hline  
  \textbf{Sequential} & 36.23 & 102.663 & 257.649 \\
  \hline
  \textbf{Pipelined} & 31.47 & 94.511 & 252.323 \\
  \hline
  \textbf{Hybrid} & 17.83 & 53.888 & 175.314  \\
  \hline
  \end{tabular}
  \end{adjustbox}
  }
\end{table}

\begin{table}[t]
\centering
\caption{Computation time hybrid architecture takes for different sizes of test images and different values of n}
\resizebox{0.7\columnwidth}{!}{
\begin{adjustbox}{max width=9 cm, max height = 3.5 cm}
\begin{tabular}{|c|c|c|c|c|}
\hline 
\multicolumn{4}{|c|}{\textbf{Computation time (in ms)}} \\
\hline
\textbf{Image Size} & \textbf{128 X 72} & \textbf{256 X 144} & \textbf{512 X 288} \\
  \hline  
  \textbf{n=1} & 31.47 & 94.511 & 252.323 \\
\hline  
  \textbf{n=2} & 25.803 & 85.9 & 245.064 \\
  \hline
  \textbf{n=4} & 21.109 & 66.183 & 212.57 \\
  \hline
  \textbf{n=8} & 17.83 & 53.888 & 175.314  \\
  \hline
  \end{tabular}
  \end{adjustbox}
  }
\end{table}

\begin{table}[!htb]
\centering
\caption{Power consumption for different Architectures for different sizes of test images}
\resizebox{0.7\columnwidth}{!}{  
\begin{adjustbox}{max width=9 cm, max height = 3.5 cm}
\begin{tabular}{|c|c|c|c|}
\hline 
\multicolumn{4}{|c|}{\textbf{Power Dissipation (in mW)}} \\
\hline
Image Size & 128 X 72 & 256 X 144 & 512 X 28 \\
  \hline  
  Sequential & 396 & 530 & 767 \\
  \hline
  Pipelined & 478 & 654 & 893 \\
  \hline
  Hybrid & 740 & 968 & 1,332 \\
  \hline
  \end{tabular}
  \end{adjustbox}
  }
\end{table}

\section{Conclusion And Future Scope}

Implementing a software algorithm on an FPGA enables us to exploit the hardware flexibility, parallelization, logical, electrical and physical advantages of it. Henceforth, it offers real time performance and economical power consumption on various mobile robotics applications. This paper presented three novel architectures of image segmentation, implemented on Virtex UltraScale XCVU190-2FLGC2104EES9847 FPGA.
The sequential design exploits the gate level control of system architecture allowing control over minute details of arithmetic design, whereas in pipelined design we exploit the hardware flexibility of pipelining provided by an FPGA. The hybrid architecture provides a full fledged parallel and pipelined implementation of the algorithm. The hybrid architecture overall provides at least 2X gain in acceleration, when compared to other implementation. Even though hybrid design has more power dissipation as compared to sequential but it is negligible compared to CPU.
\par 
Though currently we have been able to achieve only 2X speedup, this speedup is limited by the storage constraints of the device we had access to. With a better device higher acceleration can be achieved as this hybrid architecture has the potential to offer much better performance as it can execute in parallel more computations by exploiting the hardware parallelism thus allowing more number of subimages. However due to unavailability of a bigger and more expensive FPGA board, we are not able to verify the full potential of this algorithm.

\par

With the popularity of multicore programming and GPUs neural networks based image segmentation algorithms are gaining popularity. As a part of the future work, we would like to explore deep learning based segmentation algorithms like segnet and fully convoluted networks on FPGA. .
Focus would be more on maintaining the acceleration of the algorithm and quality of image segmentation without causing loss of information and keeping minimal power consumption.





%

\end{document}